\documentclass[11pt,a4paper]{article}
\usepackage[hyperref]{emnlp2020}
\usepackage{times}
\usepackage{paralist}
\usepackage{latexsym}

\usepackage{microtype}

\usepackage{graphicx, multirow}
\usepackage{todonotes}
\usepackage{enumitem}

\aclfinalcopy % Uncomment this line for the final submission
\usepackage{pgfplots,tikz}
\pgfplotsset{compat=newest}
\usetikzlibrary{intersections}
\usepackage{wasysym}
\usepackage{graphics}
\usepackage{multirow}
\usepackage{textcomp}
\usepackage{makecell}
\usepackage{amsfonts}


\title{Massive Multi-Document Summarization of Product Reviews \\ with Weak Supervision}

\author{Ori Shapira \thanks{~~Completed as part of an internship at Amazon.}\\
  Bar-Ilan University \\
  Ramat-Gan, Israel \\
  \texttt{obspp18@gmail.com} \\\And
  Ran Levy \\
  Amazon \\
  Tel-Aviv, Israel \\
  \texttt{ranlevy@amazon.com} \\}
\date{}

\begin{document}
\maketitle

\begin{abstract}
  Product reviews summarization is a type of Multi-Document Summarization (MDS) task in which the summarized document sets are often far larger than in traditional MDS (up to tens of thousands of reviews).
We highlight this difference and coin the term ``Massive Multi-Document Summarization'' (MMDS) to denote an MDS task that involves hundreds of documents or more. 
Prior work on product reviews summarization considered small samples of the reviews, mainly due to the difficulty of handling massive document sets. We show that summarizing small samples can result in loss of important information and provide misleading evaluation results. We propose a schema for summarizing a \textit{massive}
%the \textit{full}
set of reviews on top of a standard summarization algorithm. Since writing large volumes of reference summaries needed for advanced neural network models is impractical, our solution relies on weak supervision. Finally, we propose an evaluation scheme that is based on multiple crowdsourced reference summaries and aims to capture the massive
%entire
review collection. We show that an initial implementation of our schema significantly improves over several baselines in ROUGE scores, and exhibits strong coherence in a manual linguistic quality assessment.

%The task of multi-document summarization (MDS) commonly considers small document sets with no more than 30 documents. In the domain of product reviews, it is not uncommon to find thousands of reviews on a single product. We introduce the notion of ``Massive MDS'' (MMDS) where document sets are an order of magnitude larger than commonly considered. Designing and developing a supervised summarization system for this setup requires creative solutions for data collection and model training. We show that small document samples may not represent the full set faithfully, and propose a different framework that considers the entire document set, both for the summarization system and for collecting reference summaries.

%Our approach clusters the reviews to subsets and fixes each cluster's centroid as a ``summary'' to train with. This weakly-labeled training approach provides the high volume of data necessary for training a neural model. Additionally, we collect reference summaries, via crowdsourcing, by dividing up the reviews to small sets. In so, each product has several reference summaries to test its system summary against.

%We apply our approach to products with at least 100 reviews from the Amazon Customer Reviews Dataset\footnote{\url{https://s3.amazonaws.com/amazon-reviews-pds/readme.html}}. Our model improves over several baselines by atleast 14\% in ROUGE scores, and receives good results in a manual linguistic quality assessment. The product reviews' reference summaries will be publicly released.
\end{abstract}

\maketitle

\section{Introduction}
\label{sec_introduction}
Online shopping provides great convenience and flexibility for customers, however, it affects the ability to physically examine products of interest. To support the customer need for gaining familiarity with products, e-commerce websites provide a platform for customers to share their experience through online product reviews.
However, as these websites grow in popularity, so do the number of reviews, to the point that it becomes practically impossible to digest this wealth of information. Product reviews summarization aims to alleviate this problem by analyzing the entire review set and providing customers with just the right amount of information they need.

While the task of multi-document summarization (MDS) typically considers document sets with no more than 40 documents,
%\footnote{See Section \ref{sec_relatedWork} and supplementary material for a list of MDS datasets along with their document set size.}
in the domain of product reviews it is possible to find thousands of reviews on a single product. We introduce the notion of ``Massive MDS'' (MMDS) where document sets are substantially larger than commonly considered. We argue that this setup introduces new challenges that require special handling both in the system design and in the evaluation.

Several prior works on product review summarization bypassed this obstacle by restricting the task to a small sample of reviews from the entire collection, \citep[e.g.][]{angelidis2018oposum, chu2019meansum}. Small samples may not represent the full set faithfully, and systems that rely on them may neglect salient information that should be included in a summary. Another issue that arises when dealing with massive amounts of documents is the summarizer's capacity to ingest them all. Most modern summarization systems based on neural networks are limited to hundreds of words \citep{see2017pointergen, chu2019meansum, chen2018fastAbsRl}, while in the MMDS setup the summarizer may be required to process tens of thousands of words and even more.

We propose a framework that considers a \textit{massive} document set. The framework makes use of an existing summarization algorithm as an underlying component, but does not depend on its specific characteristics. In theory, any text-to-text architecture could serve as the underlying algorithm.

Our approach clusters the reviews of a single product into disjoint subsets of roughly similar size and extracts a central representative review (the medoid) from each cluster to be used as a ``weak reference summary'' of all other reviews in the cluster. We then use such \textit{(cluster, representative)} pairs to train the underlying summarization system, while meeting its text length constraint.
%documents or document-sets.
This weakly-supervised approach provides us with an unlimited pool of training examples which meets the demand of advanced neural models. Note that our weak references are more suitable for training abstractive summarizers, though an extractive system could still be trained to maximize similarity to the weak reference.
%\footnote{Our approach is designed for abstractive summarization, but could adjusted for extractive summarization.} 
%The summary generation process applies a similar clustering of the reviews with the exception that cluster representatives are not removed. The trained model is run on each cluster separately and as a result outputs multiple summaries.

The summary generation process applies a similar clustering of the reviews. The trained summarizer is run on each cluster separately, to output all the corresponding summaries.
The procedure can then be hierarchically repeated, on the output summaries, to generate a final summary that covers the massive set of reviews.

%\textbf{TODO: show that the centroid is a good approximation of human annotations, this can be done based on some of the data we already collected}.

Summarization systems are commonly evaluated against manually written reference summaries using the ROUGE \citep{lin2004rouge} family of measures. Reference summaries are written by humans,
%experts or crowd workers
after reading the documents to be summarized. In the MMDS task, this is completely infeasible for a human annotator. We overcome this limitation by, again, splitting the set of reviews to small disjoint subsets.
%borrowing ideas from our MMDS framework, with some adaptations. The set of reviews is split into small disjoint subsets.
%In order to avoid biasing the evaluation towards our solution, we do not use a clustering algorithm and instead divide the reviews randomly.
For each subset, we collect a reference summary via crowdsourcing. 
By doing so, each product has several reference summaries to test its system summary against. Note that evaluating summaries with multiple references is a common approach except that in our case, each reference is based on a different ``slice'' of the review set. 
%The reference summary dataset, named Pro-MMDS (Product MMDS) is based on 123 products with at least 100 reviews from 6 product categories in the Amazon Customer Reviews Dataset\footnote{\url{https://s3.amazonaws.com/amazon-reviews-pds/readme.html}} and will be publicly released.\footnote{Link will be provided upon acceptance.}
Our reference summary dataset is based on 123 products with at least 100 reviews taken from the Amazon Customer Reviews Dataset\footnote{\url{https://s3.amazonaws.com/amazon-reviews-pds/readme.html}}.

An implementation of our MMDS schema, 
%using straightforward
%pre-processing
%techniques and the system released by 
on top of the system released by \citet{chen2018fastAbsRl} as the underlying summarizer, significantly improves over various baselines in several ROUGE metrics,
% by at least 7.5\% to 26.5\% in several ROUGE variants,
and receives very good results, comparable to those of human written reviews, in manual linguistic quality assessments.

%\textbf{summarize contribution: framework for mmds, weak supervision, dataset}

In the next section, we report on related work, and in Section \ref{sec_motivation} we motivate our work by investigating the implications of summarizing and evaluating against small samples of product reviews.
%the importance of considering the full set of reviews when summarizing them.
Section \ref{sec_method} describes our framework for handling large collections of documents.
%Section \ref{sec_implementation} describes our implementation of the proposed schema and our annotated dataset, and 
Section \ref{sec_results} presents the experiments conducted with our implementation, as well as our MMDS dataset.

\section{Related Work}
\label{sec_relatedWork}
As MMDS is a variant of MDS,
%Our MMDS task is a variant of the more general MDS task, hence 
we start by presenting MDS in general and proceed to multi-review summarization in particular. We then provide a short survey of existing MDS datasets in order to justify the creation of a dedicated MMDS dataset.

%\paragraph{MDS Methods}

\noindent
\textbf{MDS methods.}~~~Over the years, both extractive and abstractive MDS have been approached with graph-based methods \citep[e.g.][]{erkan2004lexrank, christensen2013mds, yasunaga2017graphNeuralMds}, integer linear programming \citep[e.g.][]{bing2015mdsIlp, banerjee2015mdsIlp} and sentence or phrase ranking/selection \citep[e.g.][]{cao2015ranking, nallapati2017summarunner, fabbri2019multinews}.

Training neural networks for MDS, requires large amounts of \emph{(document set, summary)} pairs.
Recently, \citet{Liu2018wikipediaLargeScale} devised a model that generates Wikipedia articles for a given set of documents from the web. Their system processed large textual inputs by first extracting salient sentences and then feeding them into a memory optimized variant of the transformer model \citep{vaswani2017transformer}.
%Recently, \citet{Liu2018wikipediaLargeScale} devised a transformer model \citep{vaswani2017transformer} that trained on Wikipedia articles as summaries, and their corresponding reference documents as input. Their system was required to process tens of documents, some of which of considerable length. In order to overcome this hurdle they employed an extractive stage which filtered salient sentences from the full text and developed a memory optimized variant of the transformer model.
Another approach for developing MDS systems is to adapt a single-document summarization (SDS) model to MDS \citep{lebanoff2018adapting, Baumel2018QueryFA, zhang2018towards}. While the challenge of overcoming redundancy and coreference resolution
%the resolution of coreferring entities/events
is more pronounced in MDS, such adaptations leverage advancements in SDS systems.

%\paragraph{Review Summarization}
\noindent
\textbf{Review summarization.}~~~Summarizing product or service
%(e.g. restaurant, hotel, etc.)
reviews has been extensively explored both in academia and industry as e-commerce websites strive for improved customer experience and analytical insights. The most common approach is termed \emph{aspect based summarization} in which the summary is centered around a set of extracted aspects and their respective sentiment.

One of the early works, 
%One of the first significant works was
by \citet{hu2004summarizingReviews} was designed to output lists of aspects and sentiments, which is more restricted than our setup. Their system did not limit the size of the review set, nevertheless, evaluation was performed on the first $100$ reviews of only $5$ products.
%and it only covered the correctness of aspect and sentiment extraction, ignoring the summary's quality.
Other works target the summarization task, but mostly summarize small samples of reviews, and at times somewhat simplify the task by assuming aspects or seed words are provided as input \citep{Gerani2014prodReviewAbsSumm, angelidis2018oposum, yu2016reviewSumm}. Their evaluations are either ROUGE-based, on small samples of reviews, or manual pairwise summary comparisons. A variant of this manual evaluation requires evaluators to first read all reviews on a respective product, a requirement that cannot be reasonably met. This issue was raised by \citet{Gerani2014prodReviewAbsSumm} who nevertheless did not offer any remedy.

%based on this approach include e.g. \citet{Gerani2014prodReviewAbsSumm} (abstractive) and  \citet{angelidis2018oposum} (extractive) both of which consider small review samples for train/dev and for ROUGE-based evaluation.

%\citet{yu2016reviewSumm} generate summaries by running an integer linear program for salient phrase extraction. To evaluate the summaries, they carry out a manual pairwise comparison after asking the human evaluators to read all reviews on a product. However, it is fair to assume that evaluators likely read a small subset of these reviews when presented high volumes, and likely not in depth \citep{Gerani2014prodReviewAbsSumm}.

The most relevant work to ours is that of \citet{chu2019meansum} as it is an unsupervised abstractive product reviews summarizer that employs a neural encoder-decoder model. In their setup, the system works on samples of just $8$ reviews per product, and is evaluated against reference summaries based on $8$ reviews per product as well.

%\paragraph{MDS Datasets}
\noindent
\textbf{MDS datasets.}~~~
%While MDS can serve as a key component of many industrial applications and already attracts academic interest,
The main obstacle towards developing state of the art MDS models and reliably comparing between them is a shortage of large scale high-quality datasets.
The first MDS datasets originated in the DUC and TAC benchmarks\footnote{\url{https://{duc, tac}.nist.gov}}, focusing mostly on the news domain. Recently, \citet{fabbri2019multinews} released the large-scale Multi-News dataset. For Wikipedia, \citet{Liu2018wikipediaLargeScale} provide web documents with corresponding Wikipedia articles, and \citet{zopf2018hMDS} released a multilingual dataset. In the consumer reviews domain, Opinosis \citep{ganesan2010opinosis}, OpoSum \citep{angelidis2018oposum}, and a dataset by \citet{chu2019meansum} are rather small scale.
The document set sizes of the listed MDS datasets range from $2$ to $40$, 
%The MDS datasets listed contain document sets of size ranging from 2 to 40 documents,
averaging less than $10$ documents per set.
Table \ref{tab_prev_datasets_stats} presents size statistics of the aforementioned datasets in comparison with the dataset we collected as part of this work.

\begin{table}[t]
\centering
\resizebox{\columnwidth}{!}{%
\begin{tabular}{ll|c|c|c|c}
                           & \textbf{Dataset}     & \textbf{\# sets}       & \thead{\textbf{\# docs} \\ \textbf{per set}}    & \thead{\textbf{\# tokens} \\ \textbf{per doc}} & \thead{\textbf{\# tokens} \\ \textbf{per ref}}                  \\ \hline
\multirow{4}{*}{\rotatebox[origin=c]{90}{\small Reviews}}   & Opinosis   & 51            & $\dagger$  &  -             &  -                              \\
                           & MeanSum    & 200           & 8              & 70      & -                               \\
                           & Oposum     & 60            & 10             & 70      &  -                              \\
                           & MMDS (Ours) & 123           & 205            & 73            & 59 \\ \hline
\multirow{3}{*}{\rotatebox[origin=c]{90}{\small News}}      & DUC\textsuperscript{$\ddagger$} 01'-07'  & 45            & 17  & 600     & 200                      \\
                           & TAC\textsuperscript{$\ddagger$} 08'-11' & 45 & 10             & 600     & 100                            \\
                           & MultiNews   & 55K     & 3 & 700     & 260                      \\ \hline
\multirow{2}{*}{\rotatebox[origin=c]{90}{\small Wiki}} & hMDS      & 91            & 14       & 2000    & 250                      \\
                           & \citep{Liu2018wikipediaLargeScale}    & 2.3M    & 40       &    -           &  -
\end{tabular}%
}

\caption{Approximate average MDS dataset statistics. The named datasets listed are: Opinosis \citep{ganesan2010opinosis}, MeanSum \cite{chu2019meansum}, Oposum \citep{angelidis2018oposum}, DUC (\url{https://duc.nist.gov}), TAC (\url{https://tac.nist.gov}), MultiNews \citep{fabbri2019multinews}, hMDS \cite{zopf2018hMDS}.  \\ $\dagger$ Opinosis concatenates 100 sentences from different reviews. \\ $\ddagger$ In DUC and TAC datasets, values are averaged over all years.}
\label{tab_prev_datasets_stats}
\end{table}

\section{Motivation}
\label{sec_motivation}
In order to substantiate the need for MMDS, we perform preliminary analyses that demonstrate two observations. First, that products with large amounts of reviews are frequent enough to deserve special consideration, and second, that summarizing small samples of the review set may result in summaries that do not faithfully capture the salient information of the entire review set.   

We base the first observation on the statistics in Table \ref{tab_acrd_stats}. According to the table, products with more than $100$ reviews account for only $1$\% of the products in the Amazon Customer Reviews Dataset. However, their absolute number is above $200$K making it infeasible to rely on manual summaries. Furthermore, while these products represent only a small fraction of the product portfolio we argue that these are the ``interesting'' products as they are the ones customers choose to spend time on, by writing reviews. Indeed the ratio of reviews of products with more than $100$ reviews to all reviews in the dataset is approximately $0.41$. 

\begin{table}[t]
	\centering
	\resizebox{\columnwidth}{!}{%
		\begin{tabular}{l|c|c|c|c}
			\textbf{Size of} & \multicolumn{2}{c|}{\textbf{Products}} & \multicolumn{2}{c}{\textbf{Reviews}} \\ \cline{2-5}
			\textbf{Review Set} & \textbf{Count} & \textbf{Ratio} & \textbf{Count} & \textbf{Ratio} \\ \hline
%			\textbf{Review Set Size} & \textbf{\# Products} & \textbf{Product Ratio} & 
			1-9 &  19M & 0.89 & 40M & 0.25 \\
			10-99 & 2M & 0.10 & 55M & 0.34 \\
			100-999 & 200K & 0.01 & 46M & 0.28 \\
			1000-9999 & 8K & $<0.01$ & 16M & 0.10 \\
			$\geq 10000$ & 187 & $<0.01$ & 4M & 0.03 \\
		\end{tabular}%
	}
	\caption{Statistics of the Amazon Customer Reviews Dataset with respect to review set sizes.}
	\label{tab_acrd_stats}
\end{table}

As to the second observation, a good summary is expected to surface salient information from the original text(s). However, most if not all academic works on product review summarization, ignore the content of all but a few of the original texts since they are restricted to small samples of the reviews. 
We would like to measure how different sample sizes of the original texts affect information saliency. For ease of the analysis, we consider n-gram frequency as a proxy for information saliency. \citet{nenkova2006summFactors} found that high frequency words from the source texts are most agreed upon to be included in reference summaries. They reached a similar conclusion 
%when performing the analysis on content units instead of unigrams.
at the content-unit level. We thus deduce that n-gram frequencies are likely to provide a good indication for information saliency in the texts.
We measure the correlation between n-gram distributions of the entire document set and n-gram distributions of random samples of that set. If the correlation is low, we assume that the sample does not faithfully capture the information saliency of the entire document set.

%We randomly selected $210$ products from $7$ categories with a median of $195$ reviews per product. 
We randomly selected $180$ products from $6$ categories with a median of $200$ (ranging from 100 to 24K) reviews per product.
%\textsuperscript{\ref{footnote_seeSupp}}
%\footnote{See a table of stats on the data in the supplementary material.}
For each product and for each sample size, $s \in \{1,2, ..., 100\}$, we extracted $30$ samples,
%$s$ with $s$ ranging from $1$ to $100$
and measured the non-stop-word n-gram distribution for $n \in \{1,2,3\}$ on each such sample. We then measured the correlation between this distribution and the distribution of the entire set, and averaged the result across products and across the $30$ samples. Figure \ref{graph_sampleCorrelationsPearson} shows the average Pearson correlation for different sample sizes. 

%\footnote{The Pearson correlation is a stronger indicator than Spearman since a long tail of infrequent n-grams strongly skews the result. A Spearman correlation curve is available in the supplementary material.}
While samples of size $10$ to $30$ may be sufficient to capture the unigram distribution, it is clear that even with samples of size $100$, the bigram and trigram distributions still differ from those of the entire set.
%Further, humans that are given many reviews during an evaluation session cannot be expected to read and remember even $10$ reviews, which, as evident from the curve, may not be sufficient.\footnote{We also analyzed the effect of sample size on the top ranked n-gram from the full set of reviews. The results were moved to the supplementary material due to lack of space.}

%Additionally we measured how well these samples capture the top non-stop-word n-gram from the entire document set. Figure \ref{graph_sampleTopNgram} shows the percent of samples, at each sample size, that the most frequent non-stop-words n-gram from the full set is in the top-5 most frequent non-stop-word n-grams in the sample. This illustrates how well saliency is captured by the samples. As the figure shows, in the analyzed products, a sample of 10 reviews has a chance of 10\% of missing the most important unigram in its top-5 unigrams.

%We additionally measured how well samples capture saliency by estimating the likelihood that the top non-stop-word n-gram from the full review set is in the set of the top 5 n-grams of a sample.\textsuperscript{\ref{footnote_seeSupp}}
%\footnote{See details in the supplementary material.}
%the top non-stop-word n-gram from the entire document set by comparing that n-gram to the top 5 n-grams in sample sets. This illustrates how well saliency is captured by the samples. We find that a sample of 10 reviews has a chance of 10\% to miss the top unigram in its top-5 unigrams.\footnote{See supplementary material for a full graph.}

Figure \ref{graph_sampleCorrelationsSpearman} presents a similar analysis based on the Spearman correlation. We observe lower correlation than in the Pearson analysis and speculate that the Spearman variant, which compares rankings, is dominated by the long tail of low ranking n-grams. Such low frequency n-grams are not important for capturing salient information. 

\begin{figure}[h]
%\centering
%\small
%\setlength{\abovecaptionskip}{-2pt}% plus 3pt minus 5pt}

\begin{tikzpicture}%[scale =0.48,label/.style={%
   %postaction={ decorate,transform shape,
   %decoration={ markings, mark=at position .39 with \node #1;}}
   %}]

    \begin{axis}[
    axis on top,
    width=\linewidth,
        grid,
      legend style={at={(0.98,0.03)},anchor=south east,font=\small},
        height=4.5cm,
        xlabel=Sample Size,
        ylabel=Percent,
       xmin=1,   xmax=100,
    ymin=0,   ymax=1.03,
        xtick={1,10,20,30,40,50,60,70,80,90,100},
        ytick={0,0.2,...,1.0},
    ]

    \addplot[   color=black,
                %mark=triangle*,
                %mark options={solid, fill=white},
                smooth] coordinates {
        (1, 0.40177777777777784)
        (4, 0.7282222222222223)
        (7, 0.8402222222222222)
        (10, 0.8993333333333335)
        (13, 0.9366666666666666)
        (16, 0.94)
        (19, 0.9617777777777777)
        (22, 0.9651111111111114)
        (25, 0.9706666666666667)
        (28, 0.9757777777777779)
        (31, 0.9793333333333334)
        (34, 0.9826666666666665)
        (37, 0.9857777777777779)
        (40, 0.9906666666666668)
        (43, 0.9895555555555556)
        (46, 0.9884444444444443)
        (49, 0.9924444444444445)
        (52, 0.9944444444444446)
        (55, 0.9920000000000001)
        (58, 0.9933333333333333)
        (61, 0.9971111111111111)
        (64, 0.994888888888889)
        (67, 0.9957777777777779)
        (70, 0.9968888888888888)
        (73, 0.9962222222222222)
        (76, 0.996888888888889)
        (79, 0.9979999999999999)
        (82, 0.9975555555555555)
        (85, 0.9988888888888889)
        (88, 0.9986666666666667)
        (91, 0.9995555555555555)
        (94, 0.9997777777777778)
        (97, 0.9991111111111112)
        (100, 0.9995555555555555)
    };
    \addlegendentry{Unigram}
    %\addlegendentry{Uni Top5}

    % Bigram top-1 (all) vs. top-5 (samples)
    \addplot[   dashed,
                color=black,
                %draw=solid,
                %mark=triangle*,
                %mark options={solid, fill=white},
                smooth] coordinates {
        (1, 0.086)
        (4, 0.21288888888888888)
        (7, 0.3635555555555555)
        (10, 0.4522222222222221)
        (13, 0.5206666666666666)
        (16, 0.597111111111111)
        (19, 0.6393333333333334)
        (22, 0.6813333333333332)
        (25, 0.7128888888888889)
        (28, 0.7400000000000001)
        (31, 0.7679999999999999)
        (34, 0.7842222222222222)
        (37, 0.8037777777777777)
        (40, 0.8255555555555556)
        (43, 0.8424444444444444)
        (46, 0.8606666666666666)
        (49, 0.8642222222222222)
        (52, 0.8784444444444444)
        (55, 0.8884444444444444)
        (58, 0.8984444444444445)
        (61, 0.9062222222222223)
        (64, 0.924888888888889)
        (67, 0.9168888888888889)
        (70, 0.9271111111111111)
        (73, 0.9324444444444445)
        (76, 0.9380000000000002)
        (79, 0.9411111111111112)
        (82, 0.9451111111111112)
        (85, 0.9444444444444445)
        (88, 0.9460000000000001)
        (91, 0.9535555555555557)
        (94, 0.9640000000000002)
        (97, 0.9588888888888889)
        (100, 0.9620000000000001)
    };
    \addlegendentry{Bigram}
    %\addlegendentry{Bi Top5}
    
    % Trigram top-1 (all) vs. top-5 (samples)
    \addplot[   densely dotted,
                color=black,
                thick,
                %draw=solid,
                %mark=triangle*,
                %mark options={solid, fill=white},
                smooth] coordinates {
        (1, 0.02022222222222222)
        (4, 0.057777777777777775)
        (7, 0.0951111111111111)
        (10, 0.14800000000000002)
        (13, 0.1746666666666667)
        (16, 0.23755555555555555)
        (19, 0.264)
        (22, 0.2948888888888889)
        (25, 0.338)
        (28, 0.3768888888888889)
        (31, 0.4091111111111111)
        (34, 0.43600000000000005)
        (37, 0.4595555555555556)
        (40, 0.4851111111111112)
        (43, 0.5082222222222221)
        (46, 0.5206666666666666)
        (49, 0.5442222222222222)
        (52, 0.5624444444444444)
        (55, 0.581111111111111)
        (58, 0.5973333333333334)
        (61, 0.6171111111111112)
        (64, 0.6215555555555555)
        (67, 0.6451111111111112)
        (70, 0.6431111111111111)
        (73, 0.666)
        (76, 0.6933333333333334)
        (79, 0.6824444444444445)
        (82, 0.7048888888888889)
        (85, 0.7135555555555555)
        (88, 0.7288888888888889)
        (91, 0.7426666666666667)
        (94, 0.7575555555555555)
        (97, 0.7739999999999999)
        (100, 0.778888888888889)
    };
    \addlegendentry{Trigram}
    %\addlegendentry{Tri Top5}

\end{axis}

\end{tikzpicture}

\caption{The percent of samples (out of 30 samples) where the top non-stop-word n-gram in the full set of reviews is in the top-5 non-stop-word n-grams in a sample of reviews.}
%\caption{The percent of samples (out of 30 samples) where the top n-gram in the full set of reviews is in the top-1 or top-5 n-grams in a sample of reviews.}

\label{graph_sampleTopNgram}

\end{figure}
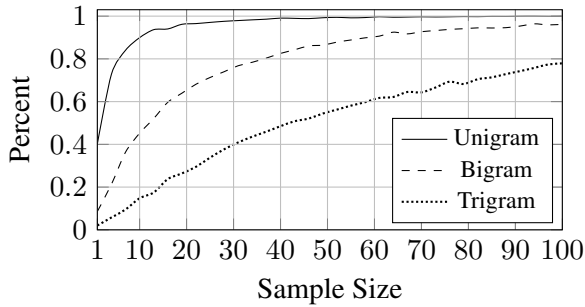

Figure \ref{graph_sampleTopNgram} shows the percent of samples, at each sample size, in which the most frequent non-stop-words n-gram from the full set is in the top-5 most frequent non-stop-word n-grams in the sample. When this condition is not met, an automatic summarization system will most likely miss out on crucial information. As the figure shows, a sample of 10 reviews has a chance of 10\% to miss the most important unigram in its top-5 unigrams.

For a qualitative impression, consider the ``Echo Dot (3rd generation)'' smart speaker that has, as of writing this paper, roughly $62K$ customer reviews on the Amazon.com website. One important aspect that is frequently mentioned in the reviews is the sound quality. The unigram \emph{sound} appears in $13K$ of the reviews and is the most frequent non-stop-word apart from \emph{love}, \emph{echo}, \emph{alexa} and \emph{great}. Sound quality is clearly a salient theme that should appear in a good summary. However, based on these numbers we can estimate that in $1$ out of $10$ samples of size $10$, the unigram \emph{sound} will not appear \textit{at all}.

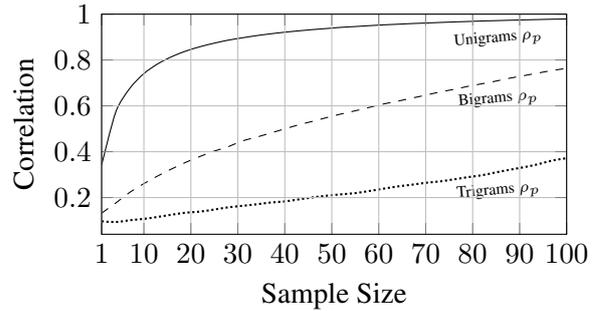
\begin{figure}[!t]
%\centering
%\small
%\setlength{\abovecaptionskip}{-2pt}% plus 3pt minus 5pt}

\begin{tikzpicture}%[scale =0.48,label/.style={%
   %postaction={ decorate,transform shape,
   %decoration={ markings, mark=at position .39 with \node #1;}}
   %}]

    \begin{axis}[
    axis on top,
    width=\linewidth,
        grid,
      legend cell align=left,
      legend columns=2,
      %legend pos=inner south east,
      %legend style={draw=none},
      legend style={at={(0.98,0.03)},anchor=south east,font=\small},
        height=4.5cm,
        xlabel=Sample Size,
        ylabel=Correlation,
       xmin=1,   xmax=100,
    ymin=0.04,   ymax=1.0,
        xtick={1,10,20,30,40,50,60,70,80,90,100},
        ytick={0,0.2,...,1.0},
    ]

    % Pearson Unigrams
    \addplot[   color=black,
                %mark=*,
                %mark options={solid, fill=white},
                smooth] coordinates {
        (1, 0.34145443952283805)
        (4, 0.5727797120718406)
        (7, 0.6763651697499761)
        (10, 0.7407817216542799)
        (13, 0.7837507242793508)
        (16, 0.8151395745706179)
        (19, 0.8393835609463157)
        (22, 0.8581341083850161)
        (25, 0.8734350670768104)
        (28, 0.8866312520873535)
        (31, 0.8967984767812737)
        (34, 0.9062213147129654)
        (37, 0.914225091333469)
        (40, 0.9212646625114902)
        (43, 0.9271218785852455)
        (46, 0.9326770123315842)
        (49, 0.9373347326948577)
        (52, 0.9424780526678339)
        (55, 0.9458412671601739)
        (58, 0.9496600645713886)
        (61, 0.9531115604184313)
        (64, 0.9559147306497919)
        (67, 0.958834119033879)
        (70, 0.9611793756971576)
        (73, 0.9636331190153029)
        (76, 0.9659185520814251)
        (79, 0.9678687552659251)
        (82, 0.9700701535565641)
        (85, 0.9715779875823609)
        (88, 0.9734210807660312)
        (91, 0.9749220389644322)
        (94, 0.9764043731986671)
        (97, 0.9779071167390462)
        (100, 0.978931333456387)
    } node[pos=0.85, rotate=5, below] {\tiny{Unigrams $\rho_p$}};

    % Pearson Bigrams
    \addplot[   dashed,
                color=black,
                %draw=solid,
                %mark=*,
                %mark options={solid, fill=white},
                smooth] coordinates {
        (1, 0.13216486518366763)
        (4, 0.1760954238887446)
        (7, 0.22289575934220066)
        (10, 0.2609226432367395)
        (13, 0.29491297290682056)
        (16, 0.32575249593306604)
        (19, 0.3544844303483377)
        (22, 0.3797469301164737)
        (25, 0.40277998320308633)
        (28, 0.42227376626666807)
        (31, 0.4474032848165078)
        (34, 0.4634891900505689)
        (37, 0.4817143982742591)
        (40, 0.4995880070319668)
        (43, 0.5177033085675478)
        (46, 0.5328901679738283)
        (49, 0.5492781441574848)
        (52, 0.5645798926885016)
        (55, 0.5784895945792115)
        (58, 0.5933835827879329)
        (61, 0.6081091955877918)
        (64, 0.6201373069776677)
        (67, 0.6335898348859403)
        (70, 0.6464980939491147)
        (73, 0.660475708658775)
        (76, 0.6717999552721813)
        (79, 0.6851646870192518)
        (82, 0.6964972205268141)
        (85, 0.7081396542921501)
        (88, 0.7210342104383487)
        (91, 0.7321836099673179)
        (94, 0.7433280878882796)
        (97, 0.7545351412006811)
        (100, 0.7633435389894172)
    } node[pos=0.85, rotate=5, below] {\tiny{Bigrams $\rho_p$}};

    % Pearson Trigrams
    \addplot[   densely dotted,
                color=black,
                thick,
                %draw=solid,
                %mark=*,
                %mark options={solid, fill=white},
                smooth] coordinates {
        (1, 0.0967573489045707)
        (4, 0.0930360016361353)
        (7, 0.10128310334409624)
        (10, 0.1070688767645036)
        (13, 0.1158019401499592)
        (16, 0.1256306104907416)
        (19, 0.13420543891212106)
        (22, 0.13885811360464312)
        (25, 0.14790958972002233)
        (28, 0.15751083244257136)
        (31, 0.16405874655822367)
        (34, 0.17063080699689329)
        (37, 0.17851243334060113)
        (40, 0.18311771801802876)
        (43, 0.19210389717873022)
        (46, 0.2008718762811085)
        (49, 0.20830568556247644)
        (52, 0.21407356017926954)
        (55, 0.2194068974683725)
        (58, 0.2287076619061678)
        (61, 0.23864268936970798)
        (64, 0.24851524824223456)
        (67, 0.2559119999980974)
        (70, 0.2644669907134517)
        (73, 0.2704565484722891)
        (76, 0.2794756877751568)
        (79, 0.28907428720376105)
        (82, 0.29742669665845767)
        (85, 0.3100500464292744)
        (88, 0.32178814967778224)
        (91, 0.3327748377087146)
        (94, 0.3457882578507721)
        (97, 0.3619600218442126)
        (100, 0.3716136412382854)
    } node[pos=0.85, rotate=5, below] {\tiny{Trigrams $\rho_p$}};

    %\legend{Unigrams $\rho_p$,Unigrams $\rho_s$,Bigrams $\rho_p$,Bigrams $\rho_s$,Trigrams $\rho_p$,Trigrams $\rho_s$}

\end{axis}

\end{tikzpicture}

\caption{Pearson ($\rho_p$) correlations of n-gram frequencies in samples of reviews vs. in the full set of reviews. 30 samples are averaged at each sample size.}
\label{graph_sampleCorrelationsPearson}

\end{figure}
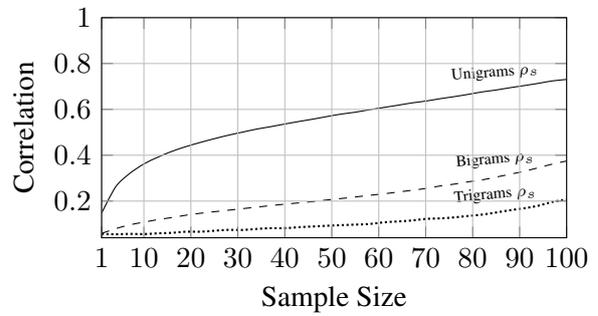
\begin{figure}[h]
%\centering
%\small
%\setlength{\abovecaptionskip}{-2pt}% plus 3pt minus 5pt}

\begin{tikzpicture}%[scale =0.48,label/.style={%
   %postaction={ decorate,transform shape,
   %decoration={ markings, mark=at position .39 with \node #1;}}
   %}]

    \begin{axis}[
    axis on top,
    width=\linewidth,
        grid,
      legend cell align=left,
      legend columns=2,
      %legend pos=inner south east,
      %legend style={draw=none},
      legend style={at={(0.98,0.03)},anchor=south east,font=\small},
        height=4.5cm,
        xlabel=Sample Size,
        ylabel=Correlation,
       xmin=1,   xmax=100,
    ymin=0.04,   ymax=1.0,
        xtick={1,10,20,30,40,50,60,70,80,90,100},
        ytick={0,0.2,...,1.0},
    ]

    % Spearman Unigrams
    \addplot[   color=black,
                %mark=triangle*,
                %mark options={solid, fill=white},
                smooth] coordinates {
        (1, 0.14743144811964873)
        (4, 0.262796620630118)
        (7, 0.32091997318806864)
        (10, 0.36201634253420695)
        (13, 0.39176836984332064)
        (16, 0.4171252380294197)
        (19, 0.43787119114407097)
        (22, 0.4557476117689367)
        (25, 0.472059908814792)
        (28, 0.48740915835922194)
        (31, 0.5007334341404132)
        (34, 0.5139509639768514)
        (37, 0.5245024133747644)
        (40, 0.5361455127543413)
        (43, 0.5471612541084208)
        (46, 0.5574860413093875)
        (49, 0.5688166488801965)
        (52, 0.5794456999496396)
        (55, 0.5876867790751583)
        (58, 0.5983414536306143)
        (61, 0.6081737141737019)
        (64, 0.6176778215307743)
        (67, 0.6280091024564144)
        (70, 0.636085724645384)
        (73, 0.6463295231357418)
        (76, 0.6553748627681258)
        (79, 0.66533746592208)
        (82, 0.6761872790601472)
        (85, 0.6845055931609797)
        (88, 0.6946597282935464)
        (91, 0.7038583225065638)
        (94, 0.7136373985937814)
        (97, 0.724229900052894)
        (100, 0.7305129993238013)
    } node[pos=0.85, rotate=5, above] {\tiny{Unigrams $\rho_s$}};

    % Spearman Bigrams
    \addplot[   dashed,
                color=black,
                %draw=solid,
                %mark=triangle*,
                %mark options={solid, fill=white},
                smooth] coordinates {
        (1, 0.058979632085487545)
        (4, 0.07971284094666241)
        (7, 0.09651467294591375)
        (10, 0.10800617072904697)
        (13, 0.1193517409633233)
        (16, 0.12824723001760244)
        (19, 0.13852402800144953)
        (22, 0.14682412625939467)
        (25, 0.15357760555411198)
        (28, 0.1602188562450642)
        (31, 0.16703175593954386)
        (34, 0.17348856627671494)
        (37, 0.18034675291329283)
        (40, 0.18639414542910288)
        (43, 0.1922576200658566)
        (46, 0.19786776856734062)
        (49, 0.2049584946791566)
        (52, 0.21112049903159144)
        (55, 0.2177130722153235)
        (58, 0.2251401950276926)
        (61, 0.23146944161335972)
        (64, 0.23915922354552574)
        (67, 0.2468908168548976)
        (70, 0.254947152078245)
        (73, 0.2656982894302503)
        (76, 0.2737659623629252)
        (79, 0.2839324083770931)
        (82, 0.2923422502941358)
        (85, 0.304794166029749)
        (88, 0.31720148692002276)
        (91, 0.33019471126591987)
        (94, 0.34560572342819273)
        (97, 0.3615176391006139)
        (100, 0.3747473003410164)
    } node[pos=0.85, rotate=5, above] {\tiny{Bigrams $\rho_s$}};

    % Spearman Trigrams
    \addplot[   densely dotted,
                color=black,
                thick,
                %draw=solid,
                %mark=triangle*,
                %mark options={solid, fill=white},
                smooth] coordinates {
        (1, 0.05708575501882931)
        (4, 0.05526630662419013)
        (7, 0.05705296498643003)
        (10, 0.05595454679244627)
        (13, 0.05953256890165227)
        (16, 0.06149631292976645)
        (19, 0.06675979870819063)
        (22, 0.06632981869019368)
        (25, 0.07071477256868662)
        (28, 0.07458536456222943)
        (31, 0.07409495444366798)
        (34, 0.07848609029391454)
        (37, 0.08193890033798379)
        (40, 0.08173599677094304)
        (43, 0.08621559843870223)
        (46, 0.08920829710620089)
        (49, 0.09246861588423629)
        (52, 0.09503322085431029)
        (55, 0.09745997616513868)
        (58, 0.09990712091328431)
        (61, 0.10731466754877242)
        (64, 0.11110061099829409)
        (67, 0.11522503095832075)
        (70, 0.12240622163270735)
        (73, 0.12440700178734886)
        (76, 0.128713773733336)
        (79, 0.13514212707448017)
        (82, 0.14107320328508996)
        (85, 0.15024147139311203)
        (88, 0.160397970692682)
        (91, 0.16884002171924287)
        (94, 0.1797778659541033)
        (97, 0.19674734323935533)
        (100, 0.20611145305841444)
    } node[pos=0.85, rotate=5, above] {\tiny{Trigrams $\rho_s$}};

    %\legend{Unigrams $\rho_p$,Unigrams $\rho_s$,Bigrams $\rho_p$,Bigrams $\rho_s$,Trigrams $\rho_p$,Trigrams $\rho_s$}

\end{axis}

\end{tikzpicture}

\caption{Spearman ($\rho_s$) correlations of n-gram frequencies in samples of reviews vs. in the full set of reviews. 30 samples are averaged at each sample size.}
\label{graph_sampleCorrelationsSpearman}

\end{figure}

Neural-based summarization systems are currently limited, in the size of texts that they process,
%both in the single-document setup as well as the multi-document setup, 
to hundreds of words, meaning that they cannot handle large review sets.
%\textbf{Do we need a table with systems and their input sizes?}
%in the range of [x-y] (see supplementary material for a list of systems and their respective input size).
A notable exception, described earlier, is the work of \citet{Liu2018wikipediaLargeScale} who were able to process up to $11K$ words.
% by devising a memory compressed attention mechanism.
However, even if all systems were able to handle massive review sets, existing evaluation methods, which are based on human judgments or human-written reference summaries, are still inherently limited to small samples of the document sets.
% and thus would fail due to similar reasoning. 
Further, humans that are given many reviews during an evaluation session cannot be expected to read and remember even $10$ reviews, which, as evident from the curve in Figure \ref{graph_sampleCorrelationsPearson}, may not be sufficient.

While it is possible to average noisy evaluation scores across many products to get a reasonable estimation,
summarization systems should aspire to work well on each product and not only on average. Furthermore, evaluation schemes that assign different weights for different products, e.g. larger weights to popular products, will have to rely on accurate evaluation at the level of single products. Interestingly, the need for automatic reviews summarization for popular products is stronger while at the same time they are more prone to the sampling bias when the sample size is fixed. 

%\textbf{deleted: Note that the sampling bias increases as the number of reviews
%to sample from
%increases; thus, the more popular a product is, the greater the sampling error becomes.
%Likewise, depending on samples of a constant small size, regardless of the number of reviews and the level of information variability, may induce a saturation shortfall and greater sampling errors \citep{israel1992sampleSize, morse1991strategies}.}
% note: morse1991strategies, in section "The Myth of Saturation"
%A small sample of reviews might satisfy a simple product, however the same sample size would not suffice for a more complex product.
%Therefore, if one would want to account for product popularity or complexity in the evaluation,
%% e.g. by increasing their weight, 
%it would require an accurate quality estimation at the level of a single product.

\section{Method}
\label{sec_method}
\begin{figure*}[!ht]
    \centering
    \includegraphics[width=1\textwidth]{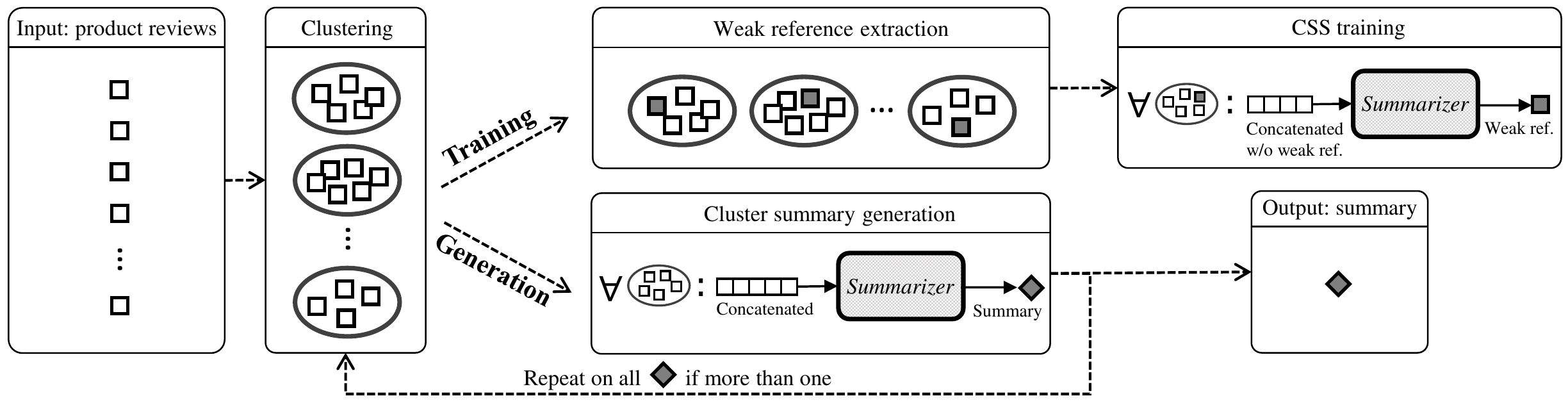}
    \caption{An illustration of our MMDS training and hierarchical cluster summarization schema.}
%Product reviews are clustered by similarity, and each cluster's weak reference is extracted. These are used for training the underlying summarization model. For summary generation, each cluster is fed into the model to get a list of summaries. The summaries are recursively clustered and summarized until a single summary is output.}
    \label{fig_schema}
\end{figure*}

Our schema is comprised of three distinct procedures for training, summary generation and evaluation, with a common theme of separating the large document set into multiple subsets and handling each of these separately. Figure \ref{fig_schema} depicts the former two procedures. In what follows we describe each procedure on a single product.
%In all three procedures we remove reviews shorter than $15$ tokens, assuming their helpfulness is negligible.
%In what follows we describe each of the processes for a single product, and assume an associated review set $R = \{r_1, r_2, ..., r_n\}$.

\subsection{Training}
\label{sec_method_training}
The training process aims to transform a set of product reviews $R = \{r_1, r_2, ..., r_n\}$ into a set of weak training examples. First, we cluster the reviews into $k$ clusters, $\mathbb{C} = \{C_1, C_2, ..., C_k\}$, such that the clusters are comparable in size and the reviews within each cluster are similar to one another. 
%are highly similar to one another.
For each cluster, $C_i$, we find a single review, $r_i^*$, with the highest similarity to all other reviews in the cluster, and denote it as the \emph{weak-reference}. If the reviews in the cluster are indeed
%highly
similar, $r_i^*$ could act as an approximate summary of all other reviews in $C_i$. 
A supervised summarization system can then be trained with data pairs $(C_i \setminus r_i^*, r_i^*)_{i \in [1, k]}$ for a practically unlimited set of products. Note that the requirement to have clusters of comparable size stems from the input size limit of the architecture we will eventually use to train on such pairs. 

The training procedure relies on three building blocks that have a large impact on the system's performance: a clustering algorithm, a similarity measure for extracting the weak-reference, and a supervised summarization system, which we term ``Cluster Summarization System" (denoted \emph{CSS}). As a proof of concept of our MMDS schema, we use the following building block implementations:

%\paragraph{Clustering.}
\noindent
\textbf{Clustering.}~~~The method used is a form of pivot clustering, constructing clusters around randomly selected pivot items, which has been shown to provide good theoretical and practical results in different settings \citep{Avigdor2016MailClustering, chierichetti2014correlationClustering, van2009pivoting}. 

As a preprocessing step, we remove from $R$ reviews shorter than $15$ tokens, assuming their helpfulness is negligible.
We initialize the unclustered review set, $U$, to the set $R$. Then, while $U$ is not empty, we randomly choose a pivot review $p$ and build a singleton cluster $C_p=\{p\}$. We then compute the ROUGE-1 $F_1$ scores between $p$ and all other reviews, and repeatedly add reviews to $C_p$, starting from the top-scoring review and moving down the scores, until $C_p$ contains \emph{min-rev} reviews, and then continue to add reviews while the accumulated text length, $\sum_{r\in{C_p}}len(r)$, is below a pre-defined threshold \emph{max-len}, where the text length is measured in sentences. In our experiments we fix \emph{max-len} to $50$ and \emph{min-rev} to $3$.
% p is set as the reference and the other review is set as the system summary

%\paragraph{Weak Reference Extraction.}
\noindent
\textbf{Weak reference extraction.}~~~Given a cluster of reviews, $C_i=\{r_i^1,...,r_i^m\}$, we measure the similarity
%define the cross-similarity 
of a review $r_i^j$ to reviews $[r_i^k]_{k=1, k\neq j}^m$ with a function $sim(r_i^j, [r_i^k])$, and define the cluster's \emph{weak-reference} as the review $r_i^*$ with the maximal $sim$ value. The training datum is then set as $(C_i \setminus r_i^*, r_i^*)$.

We experiment with different $sim$ functions. The first is the word (stem) set recall of $r_i^j$ to $[r_i^k]$, which quantifies how well $r_i^j$ covers the set of stems in $[r_i^k]$. The second is the average ROUGE-1 $F_1$ where $r_i^j$ is set as the target text and each of the reviews in $[r_i^k]$ is set as the predicted text.
%We also experimented with a similarity measure similar to the last, only with $F_2$.
While the ROUGE-1 $F_1$ variant was our first attempt, we experimented with ROUGE-1 recall,
%and ROUGE-1 $F_2$
%\footnote{The $F_2$ variant gives a stronger bias on recall, while still taking precision into account.}
%variants, 
hypothesizing that training on higher recall ``summaries'' would output longer and more informative summaries. A manual qualitative analysis revealed that output summaries were indeed longer, however they tended to contain more redundant phrases.
%Since the $F_2$ variant did not provide interesting insights, we do not report results on this variant.

%Given a cluster of reviews, $C_i=\{r_i^1,...,r_i^m\}$ and a similarity measure, $sim$, between review pairs, we define the cross-similarity of a review $r_i^j$ as $\sum_{k=1, k\neq j}^m sim(r_i^j,r_i^k)$ and define the cluster's \emph{weak-reference} as the review $r_i^*$ with maximal cross-similarity. The training datum is then set as $(C_i \setminus r_i^*, r_i^*)$. We experiment with ROUGE-1 $F_1$, ROUGE-1 recall, and ROUGE-1 $F_2$ as different similarity measures. While the $F_1$ variant was our first attempt, we experimented on recall and $F_2$\footnote{The $F_2$ variant gives a stronger bias on recall, while still taking precision into account.} variants, hypothesizing that training on higher recall ``summaries'' would output longer and more informative summaries. A manual qualitative analysis revealed that output summaries were indeed longer, however they tended to contain many redundant phrases.
%\textbf{Do we filter only based on ROUGE-1 precision? What is the best place for this filter}

In order to refrain from obtaining training examples that are difficult to train on, i.e. that would force the model to overly fabricate information in the output, we discard examples whose ``summary labels'' have too many novel unigrams. This is done by filtering out clusters where the weak-reference has a word set overlap precision of less than $0.5$.

%We filter out clusters where the weak-reference has a word set overlap precision of less than $0.5$. This prevents from obtaining a ``summary label'' that has too many novel unigrams in comparison to the reviews in its cluster. Such data is difficult to train on and requires the model to overly fabricate information in the output.

%\paragraph{Cluster Summarization System.}
\noindent
\textbf{Cluster summarization system.}~~~The CSS is a crucial element of the framework: it directly affects the final summary's quality through the quality of the cluster summaries it generates, but also indirectly by the constraints it imposes on the \emph{max-len} parameter (recall that most neural summarization systems process at most hundreds of words). After experimenting with several abstractive summarization systems, both single\footnote{The task is reduced to SDS by concatenating the reviews in a cluster to form a single input text.}
%\footnote{We reduce our cluster summarization task to single document summerization by concatenating the reviews in each cluster to form a single input text.}
and multi document, and balancing between training/generation times and manual inspection of the summaries, we found that the recent Fast Abstractive Summarization (denoted FAS) system introduced by \citet{chen2018fastAbsRl} was most promising and focused on it in our experiments.
The FAS system consists of three training phases. In the first, a sentence extraction model indicates the sentences in the input that best align to the information in the output summary. The second phase attempts to learn how to form abstractions, from the marked sentences in the first phase, to the sentences in the output. Finally, an end-to-end model utilizes the first two models to synthesize the output summary from the input.

\subsection{Summary Generation}
The summary generation process starts with a clustering phase similar to that of the training process. Given set of product reviews, $R' = \{r'_1, r'_2, ..., r'_t\}$, the reviews are clustered to $\mathbb{C}' = \{C'_1, C'_2, ..., C'_l\}$. Now, instead of converting the cluster into a training example, the trained CSS generates a cluster summary $s'_i$ for each cluster, $C'_i$. At this point we consider two alternatives to produce a single final summary. In the first, the summaries $\{s'_1, s'_2, ..., s'_l\}$ are clustered and the CSS generates summaries from the resulting clusters to produce \emph{second-level} summaries. This procedure is recursively applied until a final summary emerges. The second approach, which we refer to as the \textit{level1} approach, creates the cluster summaries as before, but then selects a single summary $s^*$ that has the highest average ROUGE-1 $F_1$ score to all other cluster summaries. The second approach aims to reduce the accumulated error when recursively applying the CSS and to prevent the final summaries from being overly generic.
%concise.

The FAS system we employ here was originally designed to summarize single documents, while we feed it a concatenation of several similar reviews or summaries. This input is expected to have higher levels of repetition.
%A concatanation of highly similar reviews, is expected to be more repetitive than a single coherent document, which FAS is designed to summarize.
Indeed, we observed that applying FAS as-is, results in somewhat repetitive summaries so we introduced a post processing step in which we measure the lemma-edit-distance between each two sentences of the summary. If the distance is above a \emph{max-edit-dist} threshold, we only keep the first sentence according to the order of appearance in the summary. In all our experiments \emph{max-edit-dist} is set to $0.7$. 

\subsection{Hyperparameters}
\label{sec_hyperparameters}
Given that our focus is on presenting a general framework for MMDS, we decided not to optimize the hyperparameters in the concrete implementation. The \emph{min-rev} parameter was set to $3$
so that one medoid could be isolated, leaving at least $2$ reviews necessary for summarizing \textit{multiple} documents.
%in order to avoid clusters which are not really MDS ($1$ or $2$ reviews).
The \emph{max-len} parameter was set to $50$ sentences as this roughly corresponds to the amount of words that FAS
%is trained to 
is designed to process. Finally, the \emph{max-edit-dist} was set to $0.7$ in order to filter cases where the repetition is very obvious.

%Practically, due to accumulating error in the recursive summary decoding procedure, the quality of the final summary dropped. I.e. information was lost on the way up, and some sentences became incoherent.
%Therefore, we also experimented with setting the final summary as the summary with the highest average ROUGE-1 $F_1$ score compared to the other summaries in the \textit{first decoding level} only.

% Should something be said about this "compromise"?

% talk about the positive/negative/neutral segmentation for first level? I think we should leave it out for now.

%The Fast Abstractive Summarization system has an improved version with repetition-avoiding reranking which takes place as an extension to the beam search. In some of our data variants this improved version did not handle the long output lengths well, in which case we only attempted to apply the simpler version without the mechanism. Still, \cite{chen2018fastAbsRl} report very good results without the extension.

\subsection{Evaluation}
\label{sec_method_evaluation}

While the field of automatic summarization has recently made a lot of progress, evaluation of such systems is still a major obstacle. Common practice relies on the ROUGE family of measures which assume that good summaries will have high n-gram overlap with human written reference summaries.
%The ROUGE measure is known to be limited \textbf{ADD citation} but is the only means of objectively comparing different systems.
A complementary approach employs human judgments for how well the system summary captures information from the original documents.

In the MMDS setup, both approaches are impractical since human annotators are not able to process so many documents in order to write a reference summary or to rate a given summary. Thus we propose to divide the reviews in a massive review set into multiple subsets, each containing an accumulated amount of up to 50 sentences, and obtain reference summaries for each subset. We believe that the clustering approach could be beneficial here as well, since it simplifies the annotator's job, however, we chose to divide the reviews randomly so as not to bias the evaluation towards our solution.
In order to evaluate a generated summary, the ROUGE score is computed for the summary against all reference summaries.
%In order to evaluate a system summary, $s^*$, ROUGE scores are computed between each reference summary and $s^*$ and the scores are averaged to produce a single ROUGE score. $ROUGE_{MMDS}(s^*)=\sum_{i=1}^kROUGE(s^*,ref_i)/k$.

Finally, since linguistic-quality evaluation does not rely on the summarized documents, coherence of MMDS summaries can be evaluated using the standard DUC linguistic quality questionnaire \citep{hoa2006ducOverview}.

%To evaluate a system summary, both a ROUGE computation and some manual assessment should be conducted. Reference summaries are required to compute ROUGE scores. As Section \ref{sec_relatedWork} mentions, \cite{chu2019meansum} crowdsourced a single summary for a small sample set of reviews of a product. In contrast, we propose to divide up the reviews to \textit{many} groups, and obtain summaries for \textit{all} groups. An average group size of $g$ reviews for a total of $n$ reviews yields $\frac{n}{g}$ groups. While the hierarchical method from above could be applied here as well, we propose to simply utilize the $\frac{n}{g}$ reference summaries
%%, or perhaps a controlled subset of them,
%as the reference summary set against which to evaluate the system summary.
% Is the following needed?
%[CITATION (Nenkova)] show that 4 reference summaries are about as effective for evaluation as 9. However a larger number could suppress some of the noise accumulated due to the crowdsourcing task.

\section{Experiments}
\label{sec_results}
\subsection{Data}
\label{ssec_refData}

We experiment with products from $6$ categories that represent different review styles, ranging from technical reviews for cameras and electronics to more prosaic reviews for books and movies (the categories are Camera, Books, Toys, Electronics, Music and DVDs).
For each product category, we randomly selected 2000 products with at least $100$ reviews from the Amazon Customer Reviews Dataset and randomly split them into $1800/100/100$ products for training, validation and test sets. Table \ref{tab_datasetStatsAllRevs} presents some statistics of the selected products.

%\input{figures/tableDatasetStatsAllRevs}

%\paragraph{Training and Validation}

\begin{table}[t]
\centering
\resizebox{\columnwidth}{!}{%
\begin{tabular}{l|c|c|c|c|c|c|c}
\multicolumn{1}{c|}{\textbf{}} & \textbf{Num}      & \multicolumn{3}{c|}{\textbf{Num Reviews}}  & \multicolumn{3}{c}{\textbf{Words/Review}}  \\
\textbf{Category}              & \textbf{Products} & \textbf{Max} & \textbf{Avg} & \textbf{Med} & \textbf{Max} & \textbf{Avg} & \textbf{Med} \\ \hline
Camera                         & 2000              & 4652         & 290          & 187          & 5877         & 67           & 33           \\
Books                          & 2000              & 8237         & 324          & 187          & 8658         & 63           & 27           \\
Electronics                    & 2000              & 15334        & 514          & 262          & 8266         & 55           & 30           \\
Music                          & 2000              & 2669         & 249          & 177          & 8693         & 86           & 37           \\
Toys                           & 2000              & 24258        & 318          & 202          & 4100         & 43           & 26           \\
DVDs                           & 2000              & 4959         & 313          & 210          & 8401         & 72           & 28           \\ \hline
All                            & 12000             & 24258        & 335          & 200          & 8693         & 63           & 30             
\end{tabular}%
}

\caption{Statistics on the full data we use as part of our analyses, training, testing and evaluation. This data is sampled from the Amazon Customer Reviews Dataset.}
\label{tab_datasetStatsAllRevs}
\end{table}
\begin{table}[t]
\centering
\resizebox{\columnwidth}{!}{%
\begin{tabular}{l|c|c|c|c|c|c|c}
\multicolumn{1}{c|}{\textbf{}} & \textbf{Num}      & \multicolumn{3}{c|}{\textbf{Num Reviews}}  & \multicolumn{3}{c}{\textbf{Words/Review}}  \\
\textbf{Category}              & \textbf{Products} & \textbf{Max} & \textbf{Avg} & \textbf{Med} & \textbf{Max} & \textbf{Avg} & \textbf{Med} \\ \hline
Camera                         & 20                & 303          & 174          & 158          & 2108         & 78           & 45           \\
Books                          & 20                & 425          & 194          & 172          & 2042         & 65           & 29           \\
Electronics                    & 22                & 717          & 235          & 169          & 1590         & 68           & 37           \\
Music                          & 20                & 445          & 203          & 196          & 2419         & 94           & 48           \\
Toys                           & 20                & 489          & 220          & 176          & 1105         & 53           & 30           \\
DVDs                           & 21                & 312          & 203          & 193          & 1964         & 82           & 33           \\ \hline
All                            & 123               & 717          & 205          & 175          & 2419         & 73           & 36             
\end{tabular}%
}

\caption{Statistics on the data in our test set, which is a subset of the data presented in Table \ref{tab_datasetStatsAllRevs}.}
\label{tab_datasetStatsTestSet}
\end{table}

\noindent
\textbf{Training and validation.}~~~The train/validation products were converted to tens of thousands of \textit{(cluster, weak-reference)} pairs.
%\footnote{Each of the weak-reference extraction variants yielded a different number of pairs due to the word overlap precision filtering performed, described in Section \ref{sec_method_training}.}
%\footnote{For example, when using the ROUGE-recall measure for weak-reference extraction, the resulting weak-references tended to be longer, and with lower precision, causing many clusters to be filtered out. In the `Camera' category, this yielded the smallest training set with about 15K datums. Otherwise the training sets contained around 40K datums.}   
Notice that thanks to the weak supervision, our framework can produce significantly larger training sets, however, this setup resulted in a reasonable tradeoff between training time and performance. 

%\paragraph{Test}
\noindent
\textbf{Test.}~~~Our evaluation scheme is based on collecting manual reference summaries for multiple subsets of each review set, as proposed in Section \ref{sec_method_evaluation}. We gathered reference summaries for about $20$ test set products, from the 100 we put aside, for each of the $6$ categories using the Figure-Eight\footnote{\url{https://www.figure-eight.com/}} crowdsourcing platform. We group reviews into \emph{annotation-sets}, with each having about $50$ sentences (but at least two reviews in a set), and present them with their star rating, and with the product title on top. The crowdsourcing task guidelines, similar to those of \citet{chu2019meansum}, are as follows:
\begin{itemize}
	\item Write a summary as if it were a review itself (e.g. to write `the screen is dark' instead of `customers thought that the screen is dark').
	\item Keep the summary length reasonably close to the average length of the presented reviews.
	\item Try to refrain from plagiarizing the original reviews by not copying more than $5$ or so consecutive words from a review.
\end{itemize}

%\footnote{See supplementary material for details.}
%, we required workers to write a summary as if it were a review itself (e.g. to write `the screen is dark' instead of `customers thought that the screen is dark'), to keep the summary length reasonably close to the average length of the presented reviews, and to try to refrain from plagiarizing the original reviews by not copying more than $5$ or so consecutive words from a review.
We automatically validated that summaries are at least $20$ tokens long.

Each annotation-set was summarized by two crowd workers. We automatically filtered out summaries that appeared vertabim more than once, summaries that were full extracts from a review, summaries with many linebreaks, and summaries that contained certain suspicious text fragments (based on manual observations on a selection of crowd-summaries).\footnote{Roughly $11.5\%$ of the annotations were filtered.} In annotation-sets for which two reference summaries remained, we heuristically selected the longer summary with the rationale that it likely contains more information.

We repeated the process on our $6$ categories, totaling $123$ products with an average of $205$ reviews per product, ranging from $100$ to $720$, and $21.75$ reference summaries per product. Table \ref{tab_datasetStatsTestSet} provides additional statistics on the test set.

\subsection{Baselines}
We compare our model to several baselines, some of them similar 
%in spirit
to those of \citet{chu2019meansum}. When generating baselines, reviews shorter than 15 and longer than 400 words were ignored.

%\paragraph{Best Recall}
\noindent
\textbf{Medoid-Recall.}~~~In section \ref{sec_method}, we hypothesize that the \emph{weak-reference} could serve as an approximate reference summary of all other cluster reviews. We can extend this hypothesis to the full review set and test whether a review with the maximal $sim$ score to all other reviews, the medoid, could be a good ``summary''. Our first baseline, which we call \emph{Medoid-Recall}, selects the review that maximizes the word (stem) set recall.
%cross similarity of the ROUGE-1 recall measure.
This measure favors reviews which cover a big portion of the review-set vocabulary.

%\paragraph{Best $F_1$}
\noindent
\textbf{Medoid-$F_1$.}~~~Here, the same technique as the previous baseline is applied, with average ROUGE-1 $F_1$ computed instead of word set recall. The intuition behind this is to mitigate the strong length bias that recall introduces, as well as to limit the amount of unique information in the selected review.

%\paragraph{Multi-Lead-1}
\noindent
\textbf{Multi-Lead-1.}~~~It is well known that the lead-$k$ technique is considered a strong single-document summary baseline in certain domains \citep{see2017pointergen}. A lead-$k$ summary merely truncates input documents after the first $k$ sentences. In the case of multiple documents, and especially in the product-reviews domain where documents are usually not very long, a parallel approach is to concatenate the first sentence from several of the shuffled documents until a certain length limit is reached. We limit our multi-lead-1 ``summary'' to $100$ tokens.
%\footnote{A product review is usually a few sentences long, while a news article can be some tens of sentences long. See supplementary material for stats.} 

%\paragraph{Cluster + Best $F_1$}
\noindent
\textbf{Cluster + Medoid-$F_1$.}~~~This baseline is a simulation of our \textit{level1} approach in which we cluster the reviews but then, instead of using the CSS to generate cluster-summaries, we extract \textit{weak-reference} reviews for the clusters (using the ROUGE-1 $F_1$ $sim$ function). Finally, we apply the Medoid-$F_1$ baseline on the resulting set of weak-references to produce the final ``summary''.

%\paragraph{Cluster + Best Recall}
\noindent
\textbf{Cluster + Medoid-Recall.}~~~This is similar to the previous baseline except that the final ``summary'' is selected out of the weak-reference set using the Medoid-Recall baseline.

\subsection{Automatic Evaluation Results}

\begin{table*}[t]
\centering
\resizebox{\textwidth}{!}{%
\begin{tabular}{ll|ll|ll|ll|ll|ll|ll}
 &  & \multicolumn{6}{c|}{\textbf{Electronics}} & \multicolumn{6}{c}{\textbf{Books}} \\ \cline{3-14}
 & \textbf{Model} & \multicolumn{2}{c|}{\textbf{ROUGE-1}} & \multicolumn{2}{c|}{\textbf{ROUGE-2}} & \multicolumn{2}{c|}{\textbf{ROUGE-L}} & \multicolumn{2}{c|}{\textbf{ROUGE-1}} & \multicolumn{2}{c|}{\textbf{ROUGE-2}} & \multicolumn{2}{c}{\textbf{ROUGE-L}} \\ \hline
\multirow{4}{*}{\rotatebox[origin=c]{90}{\small Our Variants}} & level1-F1 & \textbf{28.81} & (\textpm 1.11) & \textbf{4.77} & (\textpm 0.61) & \textbf{17.47} & (\textpm 0.8) & 25.8 & (\textpm 1.16) & \textbf{4.97} & (\textpm 0.58) & 16.48 & (\textpm 0.75) \\
 & level1-Recall & 27.82 & (\textpm 1.39) & 4.48 & (\textpm 0.6) & 17.43 & (\textpm 0.83) & \textbf{26.9} & (\textpm 0.82) & 4.45 & (\textpm 0.43) & \textbf{17.12} & (\textpm 0.44) \\
 & top-F1 & 26.19 & (\textpm 1.54) & 3.89 & (\textpm 0.57) & 15.82 & (\textpm 0.95) & 22.98 & (\textpm 1.79) & 3.85 & (\textpm 0.53) & 15.16 & (\textpm 0.98) \\
 & top-Recall & 24.15 & (\textpm 1.49) & 4.05 & (\textpm 0.48) & 15.15 & (\textpm 0.88) & 22.13 & (\textpm 1.28) & 3.74 & (\textpm 0.52) & 13.77 & (\textpm 0.72) \\ \hline
\multirow{5}{*}{\rotatebox[origin=c]{90}{\small Baselines}} & Medoid-F1 & 26.6 & (\textpm 1.14) & 3.18 & (\textpm 0.51) & 15.53 & (\textpm 0.79) & 25.43 & (\textpm 1.85) & 3.37 & (\textpm 0.52) & 15.55 & (\textpm 1.03) \\
 & Cluster + Medoid-F1 & 25.09 & (\textpm 1.34) & 2.83 & (\textpm 0.46) & 14.92 & (\textpm 0.89) & 23.19 & (\textpm 2.71) & 2.90 & (\textpm 0.63) & 14.53 & (\textpm 1.53) \\
 & Multi-Lead-1 & 23.74 & (\textpm 1.12) & 2.64 & (\textpm 0.44) & 13.65 & (\textpm 0.68) & 24.77 & (\textpm 1.31) & 2.87 & (\textpm 0.47) & 14.16 & (\textpm 0.59) \\
 & Cluster + Medoid-Recall & 18.43 & (\textpm 1.55) & 2.25 & (\textpm 0.32) & 10.58 & (\textpm 0.81) & 21.80 & (\textpm 1.49) & 3.16 & (\textpm 0.44) & 12.53 & (\textpm 1.01) \\
 & Medoid-Recall & 14.29 & (\textpm 0.64) & 1.84 & (\textpm 0.23) & 8.33 & (\textpm 0.41) & 19.19 & (\textpm 2.00) & 3.73 & (\textpm 1.46) & 11.63 & (\textpm 1.56)
\end{tabular}%
}
\caption{ROUGE $F_1$ scores on variants of our model and the baselines on two of the categories. The model variant name indicates the hierarchical level from which the output summary is taken (level-1 or top) and the the metric used for weak reference extraction. ROUGE score intervals express $\geq 95\%$ confidence.}
\label{tab_resultsRouge}
\end{table*}

% & level1-F2 & & & & & & & 19.76 & (\textpm 1.73) & 3.73 & (\textpm 0.52) & 14.71 & (\textpm 1.0) \\
% & top-F2 & & & & & & & 19.12 & (\textpm 1.33) & 3.06 & (\textpm 0.52) & 13.33 & (\textpm 0.94) \\ \hline

We consider four system variants in our automatic evaluation. The variants are created from the cross product of two implementation decisions:  (1) whether the final summary is taken from the top level of the hierarchy (top) or the first level (level1), and (2) the $sim$ function used for the \emph{weak-reference} extraction, i.e. word overlap recall or ROUGE-1 $F_1$.
% and ROUGE-1 $F_2$.

Table \ref{tab_resultsRouge} presents the ROUGE scores of our system variants and those of the baselines on the Electronics and Books categories. We first observe that applying the full summarization hierarchy (top) is almost consistently worse than choosing a medoid summary from the first level (level1). This could be explained by the fact that details are lost on the way up the hierarchy levels, causing the final summary to capture more generic common information.
%This could be explained by the fact that each summarization layer is presented with shorter inputs on average and as a result outputs shorter summaries until the final summary becomes too concise, capturing the most generic common information. 
%Another potential reason is that 
Additionally, clusters of summaries at higher levels in the summary hierarchy may contain elements with low pairwise similarity, quite different from the clusters that were used for training the CSS.

Comparing different similarity measures for the \emph{weak-reference} extraction did not lead to clear conclusions, with both ROUGE-1 $F_1$ and word set overlap recall interchangeably achieving the best result but with insignificant statistical difference.

%While the $F_1$ variant was our first attempt, we experimented on recall and $F_2$\footnote{The $F_2$ variant gives a stronger bias on recall, while still taking precision into account.} variants, hypothesizing that training on higher recall ``summaries'' would output longer and more informative summaries. A manual qualitative analysis revealed that output summaries were indeed longer, however they tended to contain many redundant phrases.
%were more repetitive, i.e. the model inserted redundant information which was found to be more salient by the system.
%, during the extraction phase in order to fill the space it was trained to use.

Our model achieves better scores than all baselines, and significantly so in most metrics and categories. It is evident that selecting a review based on high ROUGE-1 $F_1$ provides a relatively good representative review to ``summarize'' the rest of the reviews. We also find that the \textit{Medoid-Recall} baseline produces very long summaries at the expense of precision, severely weakening its ROUGE $F_1$ scores. Clustering first, simply filters out some of the longer reviews.

%As for the baselines, \textit{Multi-Lead-1} shows relatively strong results, but still overall worse than our model. Additionally, it is interesting to note that the \textit{Cluster + Best X} baselines which simulate the level1 variants of our model, are consistently below our model and even below the trivial \textit{Best Recall} baseline in most of the measurements.

We cannot perform a straight-forward comparison between our system and prior work because the MMDS setup is different by definition. However, when comparing to \citep{chu2019meansum}, we observe that our results are proportionally higher when compared to similar baselines, though on a different reviews dataset. Specifically, our best model significantly outperforms the baselines in ROUGE-2 and ROUGE-L (p-values $\leq 0.05$).
%Specifically, in most categories our best model significantly outperforms the baselines in ROUGE-1 (p-values $\leq 0.05$), and substantially so on ROUGE-2 and ROUGE-L.
%I.e., they increase their ROUGE-1 score by about $10\%$ from $26.8$ to 28.9, while we increase by 4 points from 25.4 to 29.4. In ROUGE-2 and ROUGE-L we have even better proportional improvement.

%Using the best model variant on the Camera category of our dataset, we trained and decoded separately on the Books category in the dataset. This category is different in nature as reviews tend to provide less technical details, and more narrative descriptions. In this category, on our best model variant from above, we score 26.38 ROUGE-1, 5.06 ROUGE-2, and 16.86 ROUGE-L. The multi-lead-1 baseline scores 25.05 ROUGE-1, 3.12 ROUGE-2, and 14.43 ROUGE-L. Here too, our score are proportionally higher than in \citep{chu2019meansum}.

%Results on three additional categories can be found in the supplementary material \footnote{We did not include the fourth category, Music, because the FAS summarizer ran out of memory during the summary generation step}. We find that the Camera, Electronics and Toys categories show certain resemblance, while the Books and Movies categories have separate similarities. The latter group is different in nature from the former as reviews tend to provide less technical details, and more narrative descriptions.

Tables \ref{tab_resultsRougeToysCamera} and  \ref{tab_resultsRougeMovies}, in Appendix \ref{sec_appendix}, present the results of our implementation on additional categories. We did not include the Music category because the FAS summarizer ran out of memory during the summary generation step. We find that the Camera, Electronics and Toys categories show certain resemblance, while the Books and DVDs categories have separate similarities. The latter group is different in nature from the former as reviews tend to provide less technical details, and more narrative descriptions.

\subsection{Manual Linguistic Quality Results}

We performed a manual linguistic quality assessment of the summaries from our system's best variant (level1-$F_1$) and from the \textit{Multi-Lead-1} and \textit{Medoid-$F_1$} baselines on our Electronics category test set. While it is known that these responsiveness-style evaluations are prone to weak replicability \citep{gillick2010crowdsourcing}, for the sake of completeness we report these results as well.

The five criteria evaluated are those introduced in the DUC evaluations \citep{hoa2006ducOverview}. Generally, they assess grammaticality, non-redundancy, referential clarity, focus, and structure and coherence.
%\footnote{For the exact questionnaire refer to \citep{hoa2006ducOverview}}
Crowdworkers were told to rate each criterion on a $1$-to-$5$ likert scale ($1$ is very poor and $5$ is very good), and each summary was evaluated by $5$ different workers. We used MACE
%\footnote{\url{https://www.isi.edu/publications/licensed-sw/mace/}}
\citep{hovy2013mace} to clean the crowdsourced results and improve our confidence in the final scores.

\begin{table}[t]
\resizebox{\columnwidth}{!}{%
\begin{tabular}{l|ccc}
\textbf{Criterion}      & \textbf{Ours} & \textbf{Medoid-$F_1$}   & \textbf{Multi-Lead-1}    \\ \hline
Grammaticality          & 3.73          & \textbf{4.09}         & 3.45              \\
Non-redundancy          & 3.55          & \textbf{3.91}         & 3.18              \\
Referential clarity     & 3.86          & \textbf{3.91}         & 3.59              \\
Focus                   & \textbf{3.86} & 3.32                  & 3.36              \\
Structure \& coherence  & \textbf{3.73} & 3.41                  & 3.23              \\
\end{tabular}%
}

\caption{Manual linguistic quality scores of our system (level1-F1 variant) and the \textit{Medoid- $F_1$} and \textit{Multi-Lead-1} baselines on the Electronics category.
%Notice that the \textit{Best Recall} ``summaries'' are human-written reviews, hence linguistic quality should be high.
}
\label{tab_resultsResponsiveness}
\end{table}

Table \ref{tab_resultsResponsiveness} presents the results. It is noticeable that the \textit{Multi-Lead-1} baseline is weakest, which is expected as the sentences are concatenated with complete disregard to each other. This behavior is expected to increase redundancy and weaken the flow of the narrative. The \textit{Medoid-$F_1$} baseline ``summaries'' are actual human-written reviews, hence their scores are expected to be high.
%, perhaps even providing an approximate upper bound.
Our system's results are close, and even surpass them in the focus and structure \& coherence criteria. The main takeout is that our summaries are quite readable, which is inherently on account of the underlying FAS system by \citet{chen2018fastAbsRl}.

Appendix \ref{sec_appendix} contains some summary output samples. Figure \ref{fig_summarySamplesCamera} exemplifies summaries generated by our system and the two baselines mentioned above, as well as a reference summary for the same camera lens. Figure \ref{fig_summarySamples} provides a few interesting system summaries from the DVD category and Figure \ref{fig_summarySamplesFunny} points at a few problematic system outputs.

\section{Conclusion}
\label{sec_conclusion}
MDS is a widely researched topic which traditionally assumes small document sets. However, the full potential of automatic summarization is unlocked when the document sets are so large that the average person would not be able to digest them. Specifically, in the domain of product consumer reviews, there may be hundreds, thousands and even tens of thousands of reviews for a single product. In this paper, we (1) institute \textit{massive} MDS by proposing a schema that can handle large product review sets in a weakly supervised manner, (2) collect a dataset of reference summaries of $123$ products covering the full set of reviews per product, and (3) implement an initial summarization system based on our schema, showing promising results. We hope that this framework sparks interest and subsequent research on MMDS.

For future work we would like to investigate alternative ways of clustering reviews and choosing their \emph{weak-references} in order to improve training quality. Specifically, we may look into methods capitalizing on aspect salience. Another natural extension to our work is to borrow the hierarchical approach from the summary generation procedure and apply it to generate a hierarchy of reference summaries, ending with a single reference summary or a handful of high quality summaries. Additionally, as product reviews tend to be rather short, we hypothesize that longer texts, such as in the news domain, would behave differently and require algorithmic adjustments.

%MDS is a widely researched task that has considered input document sets of rather small size, even in the domain of product consumer reviews where hundreds or thousands of reviews exist per product. In this paper, we (1) institute \textit{massive} MDS by proposing a schema that can handle large product review sets in a weak-labeled supervised manner, (2) release a dataset of reference summaries of 123 products covering the full set of reviews per product, and (3) implement an initial summarization system based on our schema, showing promising results. We hope that this framework sparks interest and subsequent research on MMDS.

%To further understand some aspects of our schema, we would be interested to generate reference summaries hierarchically, and not only at the first level. Additionally, as product reviews tend to be rather short, we suppose longer texts, such as in the news domain, would behave differently and require schematic and algorithmic adjustments.

%\begin{acks}
%Here are the acknowledgments.
%\end{acks}

%\clearpage
\bibliographystyle{acl_natbib}
\bibliography{main}

%\clearpage
\appendix

\section{Appendix}
\label{sec_appendix}
\begin{figure}[!ht]
    \small    
    \resizebox{\columnwidth}{!}{%
    \begin{tabular}{|p{\columnwidth}|} \hline \\
        %Camera - B000NP46K2 \\ 
        \normalsize
        \textbf{Product: ``Canon EF 16-35mm f/2.8L II USM Ultra Wide Angle Zoom Lens'' (Camera)} \\ \\
        \textbf{Reference Summary (crowdsourced -- one of several)} \\
        Ultra wide-angle and fast optical zoom that offers excellent peripheral performance throughout the zoom range. The lens with polarized filter included makes some incledible shots, ideal for a holiday trip very easy to transport and handle, the only defect with this device would be the price since it is a tittle high exceeding \$ 2200 \\ \\
        \textbf{Medoid-F1 Baseline} \\
        I've wanted a wide angle lens for a long time, and let me tell you this was worth the wait. While it is pricy, I don't regret paying what I did for it. It's just an amazing piece of glass. It comes with a nice pouch to protect it will pull strings at the top and can work well as a walk around lens. I use it on a Canon 5DMII and I've even done some great video shooting with it. When it's wide, it's very wide. I have not noticed the same softness that others are talking about. I've added a few photos that I shot with this to this page. \\ \\
        \textbf{Multi-Lead-1 Baseline} \\
        I purchased this lense a few months ago for landscape photography. Great Lens Period, you would have to hold one and use it. This Lens is really unbelievable. "The EF16-35mm f/2.8L II USM ultra wide angle ZOOM lens captures amazing colors and is a great walk-around lens. "No complain as to the image quality. This lens was as good as advertised! Everything you would expect from Canon L glass. \\ \\
        \textbf{System Summary (\textit{level1-F1} variant)} \\
        I have been very happy with the results. This is a great lens for the price. The lens is very sharp and the bokeh is great. Does everything i need it to do. I love this lens. %B000NP46K2_6.dec
        \\ \\
        \textbf{System Summary (\textit{level1-Recall} variant)} \\
        I have a canon rebel t3i and it is a great lens. I used this lens for my first wedding and I was amazed at the quality of the images I was looking for. I bought this lens to replace my canon ef 75-300mm lens and it was a great price. The lens is very sharp and sharp. The wide aperture is more than adequate for low light situations. I'm very happy with my purchase. Great product and would recommend to anyone.
        % B000NP46K2_12.dec
        \\ \\ \hline

    \end{tabular}%
    }
    \caption{An actual example of summaries.}
    \label{fig_summarySamplesCamera}
\end{figure}

\begin{table*}[t]
\centering
\resizebox{\textwidth}{!}{%
\begin{tabular}{ll|ll|ll|ll|ll|ll|ll}
 &  & \multicolumn{6}{c|}{\textbf{Toys}} & \multicolumn{6}{c}{\textbf{Camera}} \\ \cline{3-14}
 & \textbf{Model} & \multicolumn{2}{c|}{\textbf{ROUGE-1}} & \multicolumn{2}{c|}{\textbf{ROUGE-2}} & \multicolumn{2}{c|}{\textbf{ROUGE-L}} & \multicolumn{2}{c|}{\textbf{ROUGE-1}} & \multicolumn{2}{c|}{\textbf{ROUGE-2}} & \multicolumn{2}{c}{\textbf{ROUGE-L}} \\ \hline
\multirow{4}{*}{\rotatebox[origin=c]{90}{\small Our Variants}} & level1-F1 & \textbf{29.26} & (\textpm 1.87) & 4.64 & (\textpm 0.818) & \textbf{18.24} & (\textpm 1.08) & \textbf{29.25} & (\textpm 1.45) & 5.26 & (\textpm 0.69) & \textbf{17.74} & (\textpm 0.76) \\
 & level1-Recall & 27.57 & (\textpm 1.7) & 4.32 & (\textpm 0.73) & 17.03 & (\textpm 1.09) & 27.94 & (\textpm 1.53) & \textbf{5.36} & (\textpm 0.61) & 17.18 & (\textpm 0.74) \\
 & top-F1 & 27.43 & (\textpm 1.68) & \textbf{4.86} & (\textpm 0.681) & 17.48 & (\textpm 1.1) & 27.5 & (\textpm 1.88) & 5.06 & (\textpm 0.76) & 17.37 & (\textpm 0.94) \\
 & top-Recall & 23.92 & (\textpm 1.74) & 4.31 & (\textpm 0.62) & 15.29 & (\textpm 1.23) & 18.77 & (\textpm 1.95) & 4.13 & (\textpm 0.5) & 12.62 & (\textpm 0.91) \\ \hline
\multirow{5}{*}{\rotatebox[origin=c]{90}{\small Baselines}} & Medoid-F1 & 28.41 & (\textpm 1.44) & 3.9 & (\textpm 0.6) & 16.59 & (\textpm 1.01) & 28.27 & (\textpm 1.53) & 4.05 & (\textpm 0.65) & 16.6 & (\textpm 0.86) \\
 & Cluster + Medoid-F1 & 25.22 & (\textpm 2.84) & 3.76 & (\textpm 0.7) & 15.32 & (\textpm 1.4) & 27.33 & (\textpm 1.68) & 3.44 & (\textpm 0.51) & 15.69 & (\textpm 0.74) \\
 & Multi-Lead-1 & 24.72 & (\textpm 1.12) & 3.30 & (\textpm 0.58) & 14.36 & (\textpm 0.65) & 25.85 & (\textpm 1.57) & 3.65 & (\textpm 0.70) & 14.81 & (\textpm 0.78) \\
 & Cluster + Medoid-Recall & 19.57 & (\textpm 1.62) & 2.73 & (\textpm 0.42) & 11.37 & (\textpm 0.84) & 20.95 & (\textpm 1.87) & 2.83 & (\textpm 0.45) & 12.29 & (\textpm 1.09) \\
 & Medoid-Recall & 16.99 & (\textpm 1.37) & 2.52 & (\textpm 0.33) & 9.92 & (\textpm 0.76) & 16.17 & (\textpm 1.64) & 2.69 & (\textpm 0.52) & 9.52 & (\textpm 0.83)
\end{tabular}%
}
\caption{ROUGE $F_1$ scores on variants of our model and the baselines on the Toys and Camera categories.}
\label{tab_resultsRougeToysCamera}
\end{table*}

\begin{table}[h]
\centering
\resizebox{\columnwidth}{!}{%
\begin{tabular}{l|ll|ll|ll}
 & \multicolumn{6}{c}{\textbf{DVDs}} \\ \cline{2-7} 
 \textbf{Model} & \multicolumn{2}{c|}{\textbf{ROUGE-1}} & \multicolumn{2}{c|}{\textbf{ROUGE-2}} & \multicolumn{2}{c}{\textbf{ROUGE-L}} \\ \hline
 level1-F1 & 23.99 & (\textpm 1.6) & 4.15 & (\textpm 0.58) & 15.61 & (\textpm 0.98) \\
 level1-Recall & \textbf{26.75} & (\textpm 1.35) & \textbf{4.68} & (\textpm 0.49) & \textbf{16.37} & (\textpm 0.86) \\
 top-F1 & 25.26 & (\textpm 1.5) & 4.36 & (\textpm 0.55) & 16.16 & (\textpm 0.8) \\
 top-Recall & 20.13 & (\textpm 1.46) & 3.96 & (\textpm 0.47) & 12.59 & (\textpm 0.88) \\ \hline
 Med-F1 & 26.16 & (\textpm 1.39) & 4.14 & (\textpm 0.5) & 15.27 & (\textpm 0.71) \\
 C + Med-F1 & 24.94 & (\textpm 1.37) & 3.56 & (\textpm 0.44) & 14.75 & (\textpm 0.59) \\
 Multi-Lead-1 & 24.93 & (\textpm 1.36) & 3.41 & (\textpm 0.50) & 14.49 & (\textpm 0.77) \\
 C + Med-Rec & 21.43 & (\textpm 1.08) & 3.25 & (\textpm 0.31) & 11.85 & (\textpm 0.69) \\
 Med-Rec & 18.72 & (\textpm 1.01) & 3.11 & (\textpm 0.28) & 10.41 & (\textpm 0.62)
\end{tabular}%
}
\caption{ROUGE $F_1$ scores on variants of our model and the baselines, on the DVDs category.}
\label{tab_resultsRougeMovies}
\end{table}

\begin{figure}[h]
    \small    
    \resizebox{\columnwidth}{!}{%
    \begin{tabular}{|p{\columnwidth}|} \hline \\
        %\normalsize
        % DVDs - B00BWDFP5E_4
        \textbf{Product: ``Banshee: Season 1'' (DVDs)} \\
        Love true blood so much! The show is one of the best shows on tv. I love the fight scenes and the story line. \\ \\
        % DVDs - B003HC9JIW_1
        \textbf{Product: ``Start! Walking with Leslie Sansone 1 \& 2 Mile Walk'' (DVDs)} \\
        I have only done the 1-mile walking and I like the simplicity of the moves. I think this is a good workout for those who are looking for something to do. This is a great way to get started exercising again. \\ \\
        % DVDs - B00G15MDI0_23.dec 
        \textbf{Product: ``The Book Thief'' (DVDs)} \\
        The story is so touching and the acting is great. This is a beautiful story about a young girl in the world of nazi germany. \\
        \\ \hline
    \end{tabular}%
    }
    \caption{Interesting summaries generated by our model. In the first, notice that ``True Blood'' is from the same creator as ``Banshee''. The second summary recommends a beginner walker to acquire the DVD. Finally the third summary provides the general plot of the movie.}
    \label{fig_summarySamples}
    
	\vspace{4em}

	\small    
    \resizebox{\columnwidth}{!}{%
    \begin{tabular}{|p{\columnwidth}|} \hline \\
        %\normalsize
        % DVDs - B00DHHWXYY_17
        \textbf{Product: ``The Great Gatsby'' (DVDs)} \\
        I have read the book several times and have never read the books. This movie is a must see for the family and family. I read the book years ago and loved it. This is one of the best movies ever made . \\ \\
        % DVDs - B0042AGNB4_69
        \textbf{Product: ``Jillian Michaels: 6 Week Six-Pack'' (DVDs)} \\
        I bought this dvd for my husband and she loved it. This is a great workout for the whole family . \\ \\
        % DVDs - B00BWDFP5E_1.dec 
        \textbf{Product: ``Banshee: Season 1'' (DVDs)} \\
        I was hooked on this show. I am still waiting for the next season to come out on dvd. This is one of the best shows on tv. What a disappointment after all the hype. \\
        \\ \hline
    \end{tabular}%
    }
    \caption{Problematic summaries generated by our model. They all demonstrate the problem of self-contradiction.}
    \label{fig_summarySamplesFunny}

\end{figure}

\end{document}